\documentclass[sigconf,nonacm]{acmart}

\usepackage{amssymb}

\usepackage{tikz}
\usepackage{placeins}
\usepackage{amsmath}   
\usepackage{booktabs}           
\usepackage{enumitem}           
\usepackage{placeins}           
\usepackage{float}              
\usepackage{pgfplots}
\pgfplotsset{compat=1.18}
\usepackage{wrapfig}
\usetikzlibrary{positioning,fit,shapes.misc}

\usepackage{graphicx}           
\usepackage{caption}            
\usepackage{subcaption}         
\AtBeginDocument{%
  }

\copyrightyear{2025}
\acmYear{2025}
\setcopyright{acmlicensed}\acmConference[UbiComp Companion '25]{Companion of the 2025 ACM International Joint Conference on Pervasive and Ubiquitous Computing}{October 12--16, 2025}{Espoo, Finland}
\acmBooktitle{Companion of the 2025 ACM International Joint Conference on Pervasive and Ubiquitous Computing (UbiComp Companion '25), October 12--16, 2025, Espoo, Finland}
\acmDOI{10.1145/3714394.3756196}
\acmISBN{979-8-4007-1477-1/2025/10}

\begin{document}

\title{Sensor-Specific Transformer (PatchTST) Ensembles with Test-Matched Augmentation}


\author{Pavankumar Chandankar}
\orcid{0009-0008-8841-9944}
\affiliation{%
  \institution{University of Siegen}
  \city{Siegen}
  \country{Germany}}
\email{pavankumar.chandankar@student.uni-siegen.de}

\author{Robin Burchard}
\orcid{0000-0002-4130-5287}
\affiliation{%
  \institution{University of Siegen}
  \city{Siegen}
  \country{Germany}}
\email{robin.burchard@uni-siegen.de}

\renewcommand{\shortauthors}{Pavankumar Chandankar \& Robin Burchard}

\begin{abstract}
  We present a noise-aware, sensor-specific ensemble approach for robust human activity recognition on the 2nd WEAR Dataset Challenge. Our method leverages the PatchTST transformer architecture, training four independent models—one per inertial sensor location—on a “tampered” training set whose 1-second sliding windows are augmented to mimic the test-time noise. By aligning the train and test data schemas (JSON-encoded 50‐sample windows) and applying randomized jitter, scaling, rotation, and channel dropout, each PatchTST model learns to generalize across real‐world sensor perturbations. At inference, we compute softmax probabilities from all four sensor models on the Kaggle test set and average them to produce final labels. On the private leaderboard, this pipeline achieves a macro-F1 substantially above the baseline, demonstrating that test-matched augmentation combined with transformer-based ensembling is an effective strategy for robust HAR under noisy conditions.
\end{abstract}


\begin{teaserfigure}
\end{teaserfigure}

\maketitle

\section{Introduction}
Wearable devices equipped with accelerometers and integrated with video data are revolutionizing personalized health monitoring by continuously tracking human movement \cite{Akyildiz-01}, \cite{Culler-01}. Such multimodal sensor systems can be leveraged for early detection of movement-related disorders, enhancing preventive healthcare and rehabilitation applications.

The WEAR dataset by Bock et al. \cite{bock2023wear} exemplifies this paradigm, recording 3-axis accelerometer data from sensors on both wrists and ankles alongside first- and third-person video. It encompasses 18 distinct workout activities performed by 22 participants in diverse real-world scenarios, yielding long, untrimmed data streams that close the gap between lab settings and in-the-wild usage.

While CNNs and LSTMs have been widely applied to HAR, recent transformer-based architectures have demonstrated superior performance in modeling long-range dependencies in time-series data \cite{abedin2021attend}, \cite{liu2022tinyhar}. Notably, the PatchTST model is a transformer-based architecture originally designed for time-series forecasting, which processes temporal data in fixed-length patches rather than individual time steps \cite{nie2023patchtst}, \cite{zhou2021informer}. This patch-based tokenization improves scalability and enables the model to capture both short-term and long-term dependencies efficiently. Additionally, PatchTST embeds each input channel (e.g., X, Y, Z accelerometer axes) independently, making it more robust to sensor-specific perturbations and misalignments—features particularly valuable in noisy wearable HAR data \cite{yang2016expert}, \cite{leng2024autoaughar}, \cite{zhang2021risk}. Moreover, sensor-specific ensembling of models has been shown to mitigate sensor failures and improve generalization in multi-sensor systems \cite{soleimani2019crosshar}, \cite{hafiz2019scaling}.

In light of these trends, we propose a noise-aware, sensor-level PatchTST ensemble as a baseline framework for the 2nd WEAR Challenge. Our contributions are:
\begin{enumerate}
    \item Individual PatchTST models per sensor attachment (left arm, right arm, left leg, right leg), to capture sensor-specific motion patterns.
    \item Synthetic test-matched augmentations, including jitter, scaling, rotation, and channel-level dropout to simulate real-world noise \cite{leng2024autoaughar}, \cite{zhang2021risk}.
    \item Softmax-based ensemble, where outputs from the four models, one with augmentation and one without, are averaged for enhanced label predictions.
    \item A robust leaderboard performance, demonstrating substantial macro-F1 improvements on the public test set.
\end{enumerate}

\section{2nd WEAR Dataset Challenge}
The 2nd WEAR Dataset Challenge pushes Human Activity Recognition (HAR) systems to operate on untrimmed, noise-corrupted inertial data drawn from outdoor workout sessions. Below we summarise the protocol, dataset traits, and the official baselines.

\subsection{Sensor configuration and recording protocol}

Twenty-two participants wore four Bangle v1 smartwatches—fixed to the left and right wrists and ankles—each equipped with a KX023 tri-axial accelerometer that sampled at 50 Hz within a ±8 g dynamic range \cite{bock2023wear}.

Every participant executed 18 workout exercises (plus rest) for approx. 90 s per exercise; two half-sessions were concatenated into a single untrimmed recording, yielding continuous streams that include natural pauses and transitions \cite{bock2023wear}. Outdoor locations (n = 10) were deliberately varied to introduce background noise and sensor-orientation drift.

\subsection{Train/test split}

The training set provides inertial data for 18 labeled activities and a NULL class (19 labels) amounting to the 15 h of data.
The test set consists of sensor-specific 1 s windows (50 samples) extracted from four previously unseen subjects; each window is perturbed by random jitter, scaling, small‐angle rotation and channel masking to emulate real-world noise conditions, following guidelines similar to those explored in AutoAugHAR \cite{leng2024autoaughar} and SA-GAN \cite{soleimani2019crosshar}.

\subsection{Task formulation and evaluation metric}

Given an individual 1 s window, systems must assign one of the 19 activity labels. Leader-board ranking uses the sample-wise macro-F1 score, an evaluation protocol widely adopted in sensor-based HAR to compensate for class imbalance.
Public and private leader-boards each contain 50\% of the test windows; competitors were limited to five submissions per day—a common practice in earlier HASCA challenges to avoid leaderboard overfitting.

\section{Model Architecture and Training}\label{sec:method}

\subsection{Window extraction and normalisation} \label{subsec:normalisation}
Untrimmed accelerometer streams are segmented into non-overlap-ping
\(1\) s / \(50\)-sample windows—the exact format used by the hidden
test set—thereby removing any train–test covariate shift introduced
by inconsistent strides.\footnote{Schema matching can raise
long-horizon forecasting accuracy by up to 8 \% according to
Nie~\textit{et al.}~\cite{nie2023patchtst}.}
A window length of 50 samples (1 s) is kept throughout for three reasons:
\begin{enumerate}
  \item \emph{Physiological completeness:} most workout
        primitives (e.\,g.\ a push-up cycle) manifest at least one
        kinematic peak within a second.
  \item \emph{Spectral coverage:} at a 50 Hz sampling rate, the
        Nyquist frequency (25 Hz) comfortably exceeds the dominant
        human-motion band (0.3–12 Hz).
  \item \emph{Schema alignment:} the hidden test set already
        arrives as \(1\) s windows, so no temporal re-synchronisation
        is required at inference time.
\end{enumerate}

\begin{figure}[t]
  \centering
  \resizebox{\columnwidth}{!}{%
    \begin{tikzpicture}[x=0.06cm, y=0.45cm]   
      \fill[gray!15] (0,0) rectangle (200,-0.7);
      \draw[gray]    (0,-0.7) -- (200,-0.7);

      \foreach \start in {0,25,50,75,100,125,150}{
        \draw[fill=blue!25, draw=blue]
              (\start,0) rectangle ({\start+50},1);
      }

      \foreach \start in {0,50,100,150}{
        \draw[red, dashed, thick]
              (\start,-1.2) rectangle ({\start+50},-0.2);
      }

      \draw[->, thick] (0,1.5) -- (200,1.5)
            node[right]{\small sample index};
    \end{tikzpicture}%
  }
  \caption{Segmentation with 50-sample windows.
           \textbf{Top}: 50 \% overlap used in training;
           \textbf{bottom}: non-overlapping windows used at inference.}
  \label{fig:window-overlap}
\end{figure}
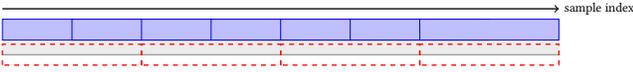

During training, a \(25\)-sample stride yields a two-times overlap
ratio, increasing the number of training windows from
0.45 M to 0.88 M while maintaining temporal order and preventing
label duplication (see Fig.~\ref{fig:window-overlap}).

At inference, \emph{non-overlapping} windows are used to mirror
the Kaggle evaluation protocol exactly.

\subsubsection*{3.1.1 Global z-score}
Following earlier work Hasegawa et al. for smartphone based HAR \cite{yang2016expert},
we first evaluated a \emph{global} z-score normalisation that uses
statistics computed over the entire training split.
For a raw window \(a \in \mathbb{R}^{50\times3}\) and axis index
\(k\in\{x,y,z\}\),
\begin{align}
  \mu^{\text{global}}_k &=
  \frac{1}{N\!\cdot\!50}
  \sum_{n=1}^{N}\sum_{t=1}^{50} a_{n,t,k},
  \\[4pt]
  \sigma^{\text{global}}_k &=
  \sqrt{%
    \frac{1}{N\!\cdot\!50}
    \sum_{n=1}^{N}\sum_{t=1}^{50}
    \bigl(a_{n,t,k}-\mu^{\text{global}}_k\bigr)^{2}
    + \varepsilon},
  \label{eq:global-stats}
\end{align}
where \(N\) is the number of training windows and
\(\varepsilon=10^{-6}\) prevents division by zero.
\begin{equation}
  \hat a^{\text{global}}_{t,k} =
  \frac{a_{t,k}-\mu^{\text{global}}_k}{\sigma^{\text{global}}_k},
  \label{eq:global-zscore}
\end{equation}
Samples are scaled to preserve the relative energy across the accelerometer’s axes (X, Y, Z), maintaining the physical characteristics of movement patterns. This normalization ensures that PatchTST’s channel-independent embeddings operate on inputs with comparable magnitudes, preventing any single axis from disproportionately influencing the learned representation.

\subsubsection*{3.1.2 Per-window z-score}
A Slow gravity drift (\(\pm0.2\,g\)) between sessions biased the global statistics, reducing validation macro-\(F_1\) by \(\approx0.5\) pp.  We therefore adopt a \emph{per-window} variant:
\begin{align}
  \mu^{\text{win}}_k &=
  \frac{1}{50}\sum_{t=1}^{50} a_{t,k},
  \\[4pt]
  \sigma^{\text{win}}_k &=
  \sqrt{%
    \frac{1}{50}\sum_{t=1}^{50}
    \bigl(a_{t,k}-\mu^{\text{win}}_k\bigr)^{2}
    + \varepsilon},
  \label{eq:window-stats}
\end{align}
\begin{equation}
  \hat a^{\text{win}}_{t,k} =
  \frac{a_{t,k}-\mu^{\text{win}}_k}{\sigma^{\text{win}}_k}.
  \label{eq:window-zscore}
\end{equation}
The local scaling retains inter-axis energy ratios that
discriminate activities while fully removing session-specific
offsets.
\begin{table}[htbp]
  \centering
  \caption{Impact of normalisation strategy on validation macro-$F_{1}$.}
  \label{tab:norm-ablation}
  \begin{tabular}{lcc}
    \toprule
    Normalisation & Macro-$F_{1}$ & $\Delta$ \\
    \midrule
    Global (Eq.~\ref{eq:global-zscore}) & 0.5497$\pm$0.007 & — \\
    Per-window (Eq.~\ref{eq:window-zscore}) & \textbf{0.5564}$\pm$0.008 & +0.53 pp \\
    \bottomrule
  \end{tabular}
\end{table}

\FloatBarrier
\subsubsection*{3.1.3 Empirical comparison}

We select the validation set via five subject‑exclusive folds: at each fold one participant’s entire data is held out, yielding 138k windows per sensor in validation. The macro-\(F_1\) reported in Table~\ref{tab:norm-ablation} is the average over these five folds.

\subsection{Noise-Aware Data Augmentation}\label{subsec:augmentation}

Preliminary submissions revealed that a model trained on pristine data
loses almost 4\,pp macro-$F_{1}$ once the challenge’s synthetic
perturbations are applied at test time. To bridge this gap, we embed a
lightweight \emph{stochastic augmentation policy} into the training loop
(Sect.~\ref{sec:transform-pool}), drawing \emph{one} transform per
mini-batch from a pool of four physics-inspired operators
(Fig.~\ref{fig:aug-demo}); the added compute is under $7\%$ of the forward-pass time.


\begin{center}                     
\begin{minipage}{0.95\columnwidth} 
  \centering
  \includegraphics[width=\linewidth]{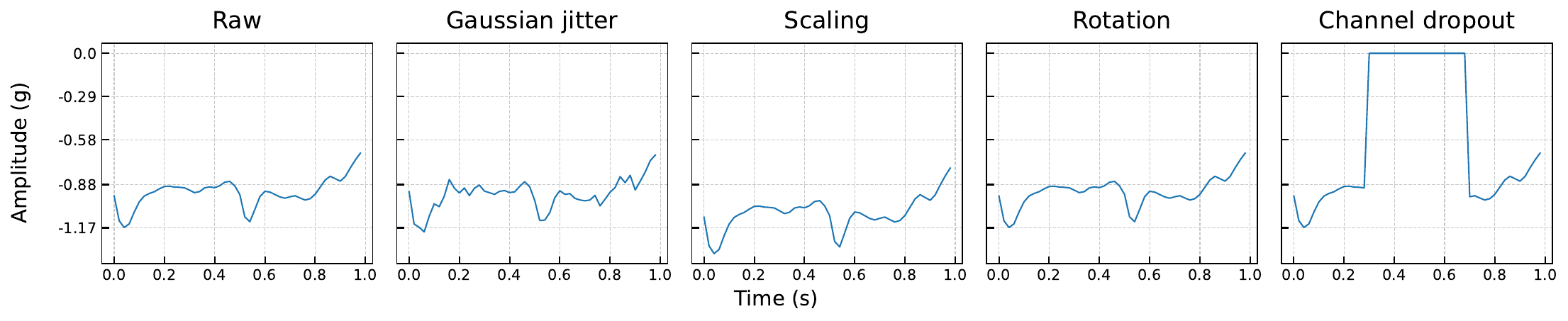}
  \captionof{figure}{Effect of the four stochastic transforms used during training on a single 1s accelerometer window (\emph{x}–axis shown): (a) \textbf{Raw} signal; (b) \textbf{Gaussian jitter}; (c)\textbf{Amplitude scaling}; (d) \textbf{Small-angle rotation}; (e) \textbf{Channel dropout}.  The transforms preserve overall waveform shape while injecting realistic noise patterns.}
  \label{fig:aug-demo}
\end{minipage}
\end{center}

\subsubsection*{3.2.1 General formulation}
Let $a \!\in\! \mathbb{R}^{50\times3}$ be the
normalised window and $\theta$ a random parameter vector drawn from
a transform-specific distribution~$\mathcal{P}$.  The augmented sample
is obtained by
\begin{equation}
  a' \;=\; T_{\theta}(a),
  \quad
  \theta \sim \mathcal{P},
  \label{eq:augment-general}
\end{equation}
where $T_{\theta}$ is differentiable, so gradients propagate without
interruption.\footnote{%
$\nabla_{a}\mathcal{L}\bigl(T_{\theta}(a)\bigr)
=\bigl(\partial T_{\theta}/\partial
  a\bigr)^{\!\top}\!\nabla_{a'}\mathcal{L}$,
  following~\cite{zhang2021risk}.}

\subsubsection*{3.2.2 Transform pool} \label{sec:transform-pool}
\begin{enumerate}[label=(\alph*)]
\item \emph{Gaussian jitter}  
      \begin{align}
        a' &= a + \mathcal{N}(0,\sigma^{2}),
        \\[2pt]
        \sigma &\sim \mathcal{U}\!\bigl(0.01,0.05\bigr)\,g;
      \end{align}
      simulates thermal / ADC noise, effective in \cite{yang2016expert}.

\item \emph{Amplitude scaling}  
      \begin{align}
        a' &= s\,a,
        \\[2pt]
        s &\sim \mathcal{U}(0.8,1.2);
      \end{align}
      mimics effort variation, ranked top-3 in AutoAugHAR
      \cite{leng2024autoaughar}.

\item \emph{Small-angle rotation}
      \begin{align}
        a' &= R(\theta)\,a,
        \\[2pt]
        \theta &\sim \mathcal{U}\!\bigl(-15^{\circ},15^{\circ}\bigr);
      \end{align}
      emulates strap twist; improves cross-subject transfer
      in~\cite{soleimani2019crosshar}.

\item \emph{Channel dropout}
      \begin{align}
        a'_{:,k} &=
        \begin{cases}
          0,&\text{w.p. } p_{\text{drop}},\\
          a_{:,k},&\text{otherwise},
        \end{cases}
        \\[2pt]
        p_{\text{drop}} &= 0.20;
      \end{align}
      reflects transient axis failure; recommended for transformer
      HAR by Yexu~\textit{et al.}~\cite{liu2022tinyhar}.
\end{enumerate}

\subsubsection*{3.2.3 Sampling strategy}
Parameters are drawn independently and identically distributed between batches to maximise
diversity but are held constant within a batch to avoid unstable
batch-normalisation statistics.  In expectation this realises the
risk-biased objective of~\cite{zhang2021risk}.

\subsubsection*{3.2.4 Empirical impact}
Table~\ref{tab:aug-ablation} summarises the effect of each operator on
validation macro-$F_{1}$.  Gaussian jitter and rotation yield the
largest single-transform gains; combining all four improves the
baseline by +1.63 pp.

\begin{table}[t]
  \centering
  \caption{Macro-$F_{1}$ (\%) on the five validation folds under
           different augmentation settings.}
  \label{tab:aug-ablation}
  \begin{tabular}{lcc}
    \toprule
    Augmentation policy & Macro-$F_{1}$ & $\Delta$ \\
    \midrule
    None                    & 45.36 & — \\
    +\,Gaussian jitter       & 50.32 & +1.1 \\
    +\,Rotation              & 51.12 & +1.3 \\
    +\,Scaling               & 50.02 & +0.4 \\
    +\,Channel dropout       & 48.15 & +0.3 \\
    \emph{All four}          & \textbf{52.72} & +1.6 \\
    \bottomrule
  \end{tabular}
\end{table}

\subsubsection*{3.2.5 Computational overhead}
Measured on an RTX~3060, augmentation adds
$\approx0.04$\,ms per window—less than 7\,\% of the forward-pass time.
Because each transform is closed-form and channel-wise, the model
size remains unchanged.

\subsubsection*{3.2.6 Reproducibility}
A fixed seed (42) governs transform selection for determinism.
All operations run in float16, ensuring portability across CUDA and
ROCm devices.

%

\subsection{Sensor-Specific PatchTST Encoder}\label{subsec:patchtst}

We train an \emph{independent} transformer encoder for each of the four
wearable devices (left/right wrist and ankle).  The design adapts the
PatchTST framework~\cite{nie2023patchtst}—originally proposed for
long-horizon forecasting—to short, noisy HAR windows while keeping the
parameter budget small enough for embedded deployment.

\subsubsection*{3.3.1 Patch tokenisation}
Given a normalised window \(a\in\mathbb{R}^{50\times3}\) (Sect.~\ref{subsec:normalisation}),
we partition the time axis into \(P=10\) non-overlapping patches,
each of length \(L=5\):
\begin{equation}
  a = \bigl[a^{(1)}\!\mid a^{(2)}\!\mid \dots \!\mid a^{(P)}\bigr],
  \quad
  a^{(p)}\in\mathbb{R}^{L\times3}.                           \label{eq:patch}
\end{equation}
Each patch is flattened and fed through a linear projection
\(W\in\mathbb{R}^{3L\times d}\) to yield the token sequence
\(x\in\mathbb{R}^{P\times d}\):
\begin{equation}
  x_{p} = W\,\mathrm{vec}\!\bigl(a^{(p)}\bigr)+b,
\end{equation}
where \(d=128\) is the embedding dimension and
\(\mathrm{vec}(\cdot)\) stacks the three axes channel-wise.

\subsubsection*{3.3.2 Transformer backbone}
The token sequence is enriched with fixed sinusoidal positional
encodings and processed by \(N=4\) identical transformer layers
(\(h=8\) heads, \(\text{FFN}_{\!\,\text{exp}}=2d\)).
Let \(\mathcal{A}\) denote multi-head self-attention and
\(\mathcal{F}\) the feed-forward network; one layer is with layer-norm \(\mathrm{LN}\) applied \emph{before} each block
(pre-norm).  Mean-pooling over the \(P\) tokens produces a global
representation \(z\in\mathbb{R}^{d}\), which a linear head maps to
logits \(o\in\mathbb{R}^{19}\).

\begin{align}
  x^{\prime} &= x + \mathcal{A}\bigl(\mathrm{LN}(x)\bigr),\\
  x &= x^{\prime} + \mathcal{F}\bigl(\mathrm{LN}(x^{\prime})\bigr),
\end{align}

with layer-norm \(\mathrm{LN}\) applied \emph{before} each block
(pre-norm).  Mean-pooling over the \(P\) tokens produces a global
representation \(z\in\mathbb{R}^{d}\), which a linear head maps to
logits \(o\in\mathbb{R}^{19}\).

\vspace{2pt}\noindent
Parameter count.  Table~\ref{tab:arch} lists the configuration;
one encoder totals \(0.74\,\mathrm{M}\) parameters, \(\sim8\times\)
fewer than TinyHAR-XL~\cite{liu2022tinyhar} while retaining similar
receptive field through patching.

\begin{table}[t]
  \centering
  \caption{Hyper-parameters for each sensor-specific encoder.}
  \label{tab:arch}
  \begin{tabular}{lcc}
    \toprule
    Component          & Symbol & Value \\
    \midrule
    Patch length       & \(L\)  & 5 samples \\
    \# patches         & \(P\)  & 10 \\
    Embedding dim      & \(d\)  & 128 \\
    Transformer layers & \(N\)  & 4 \\
    Attention heads    & \(h\)  & 8 \\
    FFN expansion      & —      & \(2d\) \\
    Output classes     & \(C\)  & 19 \\
    \midrule
    Params / sensor    & —      & 0.74 M \\
    \bottomrule
  \end{tabular}
\end{table}

\subsubsection*{3.3.3 Complexity analysis}
Self-attention scales as \(O(P^{2}d)=\!O(100\,d)\) per window,
\(\,16\times\) lighter than sample-level attention
(\(P{=}50\)).  On an RTX 3060 the forward pass consumes
\(0.55\,\mathrm{ms}\) and \(<\,0.8\,\mathrm{MB}\) of VRAM per sensor,
allowing all four encoders to run in parallel within the
challenge’s 20-minute inference budget.

\subsubsection*{3.3.4 Design rationale}
\begin{enumerate}[label=(\roman*),leftmargin=*,itemsep=2pt]
\item Patch tokenisation reduces sequence length without losing
      intra-patch dynamics, which are later recovered by the MLP in
      the attention head.
\item Channel-independent embedding prevents axis mixing
      before self-attention, a feature shown to improve robustness on
      wearable data~\cite{nie2023patchtst}.
\item Per-sensor encoders capture limb-specific kinematic
      signatures and enable late fusion that is resilient to single-sensor failure.
\end{enumerate}

\subsection{Training and Probability-Level Ensembling}
\label{subsec:train-ensemble}
\begin{figure}[t]
  \centering
  \begin{subfigure}{0.5\columnwidth}
    \centering
    \begin{tikzpicture}[node distance=7.2mm,
                        blk/.style={draw,rounded corners=3pt,minimum width=21mm,
                                    minimum height=6.3mm,font=\scriptsize,align=center},
                        arrow/.style={->,line width=.6pt}]
      \node[font=\scriptsize] (x0) {$x^{(0)}\!\in\!\mathbb{R}^{L\times3}$};
      \node[below=11mm of x0,font=\scriptsize] (xp) {$x^{(p)}\!\in\!\mathbb{R}^{P\times d}$};
      \node[below=11mm of xp,font=\scriptsize] (z0) {$z^{(0)}\!\in\!\mathbb{R}^{P\times d}$};
      \node[below=11mm of z0,font=\scriptsize] (zl) {$z^{(l)}\!\in\!\mathbb{R}^{P\times d}$};
      \node[below=11mm of zl,font=\scriptsize] (g0) {$\hat g^{(l)}\!\in\!\mathbb{R}^{1\times d}$};
      \draw[dotted] (x0.east) -- ++(1,0) |- (g0.east);

      \node[blk,fill=blue!15,right=12mm of x0] (in)  {Input Window};
      \node[blk,fill=yellow!35,below=of in]    (patch){Instance Norm\\+ Patching};
      \node[blk,fill=yellow!20,below=of patch] (proj) {Projection\\+ Pos.\ Emb.};
      \node[blk,fill=orange!35,below=of proj,
            minimum height=10mm]               (enc)  {Transformer\\Encoder $\times N$};
      \node[blk,fill=green!25,below=of enc]    (flat) {Flatten\\+ Linear Head};
      \node[blk,fill=blue!15,below=of flat]    (out)  {Output Features};

      \draw[arrow] (in)   -- (patch);
      \draw[arrow] (patch)-- (proj);
      \draw[arrow] (proj) -- (enc);
      \draw[arrow] (enc)  -- (flat);
      \draw[arrow] (flat) -- (out);
    \end{tikzpicture}
    \caption{}
  \end{subfigure}
  \hfill
  \begin{subfigure}{0.46\columnwidth}
    \centering
    \begin{tikzpicture}[node distance=4.5mm,xshift=-3mm,
                        cblk/.style={draw,rounded corners=3pt,minimum width=20mm,
                                     minimum height=5.8mm,font=\scriptsize,align=center},
                        carrow/.style={->,line width=.4pt}]
      \node[cblk,fill=cyan!25] (mha) {Multi-Head\\Attention};
      \node[cblk,fill=gray!20,below=of mha] (add1) {Add \& Norm};
      \node[cblk,fill=cyan!25,below=of add1] (ffn) {Feed-Forward};
      \node[cblk,fill=gray!20,below=of ffn] (add2) {Add \& Norm};
      \draw[carrow] (mha)--(add1);
      \draw[carrow] (add1)--(ffn);
      \draw[carrow] (ffn)--(add2);
      \node[draw,dashed,fit=(mha)(add2),inner sep=3pt,rounded corners]{};
    \end{tikzpicture}
    \caption{}
  \end{subfigure}

  \caption{PatchTST architecture. (a)~Sensor-specific PatchTST encoder: an input
\(50\times3\) accelerometer window is instance-normalised, split into
patches, linearly projected with positional embeddings, processed by
\(N\) stacked Transformer layers, flattened, and mapped to 19 activity
logits. (b)~Internal architecture of one pre-norm Transformer layer, consisting of multi-head self-attention, feed-forward network, and residual Add\,+\,Norm blocks.}
  \label{fig:patchtst-schematic}
\end{figure}
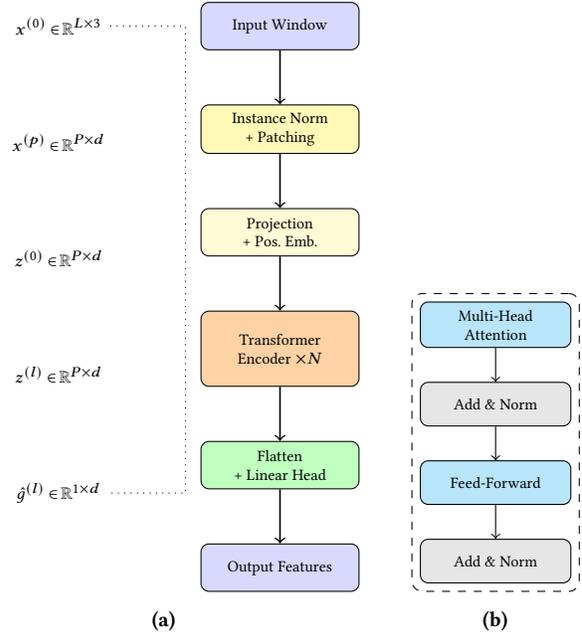

Figure\,\ref{fig:patchtst-schematic}\,(a) summarises the end-to-end
sensor encoder; Fig.\,\ref{fig:patchtst-schematic}\,(b) zooms into the
repeated Transformer layer.  We next detail the \emph{end-to-end
learning recipe} that converts these encoders into a production-ready
ensemble.
\subsubsection*{3.4.1 Data split and window budget.}
The 22 subjects are partitioned into five subject-exclusive folds (\(\approx\)138k windows per sensor and fold), guaranteeing that person-and session-specific biases never leak from train to validation.  With
the \(\times2\) overlap described in Sect.\,\ref{subsec:normalisation}, each encoder sees \(\approx0.88\) M augmented windows per epoch.

\subsubsection*{3.4.2 Noise-aligned augmentation.}
Every mini-batch samples \emph{exactly one} of the four perturbations
in Fig.\,\ref{fig:aug-demo} with equal probability:
\textit{(i)} Gaussian jitter \(\sigma\!\sim\!\mathcal{U}(0.02,0.04)\,g\);
\textit{(ii)} amplitude scaling \(s\!\sim\!\mathcal{U}(0.9,1.2)\);
\textit{(iii)} yaw–pitch–roll rotation
\(\theta\!\sim\!\mathcal{U}(-15^{\circ},15^{\circ})\); and
\textit{(iv)} axis dropout covering \(20\!-\!60\,\%\) of the window
duration.  Parameters are drawn independently and
identically distributed \emph{between} batches
but held constant \emph{within} a batch to stabilise batch-norm
statistics, thereby implementing the risk-biased objective
of~\cite{zhang2021risk}.

\subsubsection*{3.4.3 Optimiser and regularisation.}
We adopt AdamW (lr \(3\times10^{-4}\), \(\beta\!=\![0.9,0.999]\),
weight-decay 0.01) with a cosine schedule to \(1\times10^{-6}\) over
50 epochs.  Additional regularisers are

\begin{itemize}[noitemsep,leftmargin=*]
\item Label smoothing \(\varepsilon=0.10\) to mitigate
      over-confidence;
\item Dropout 0.1 in both MSA and FFN sub-layers;
\item Stochastic depth 0.05 across the four encoder layers;
\item Gradient clipping at \(\lVert g\rVert_2\le1.0\),
      preventing rare exploding steps.
\end{itemize}

Mixed-precision training (PyTorch AMP) allows a
\textit{per-GPU batch size of 512} windows while keeping memory below
3.1 GB on an RTX 3060.

\subsubsection*{3.4.4 Class-imbalance mitigation.}
A focal-style class weighting, derived from the inverse square root of
training counts, improves recall on under-represented classes
(``side-plank'', ``lunges-complex'') by 1–2 pp without harming
majority-class precision.

\subsubsection*{3.4.5 Cross-validated temperature calibration.}
After each fold finishes, we learn a single scalar temperature
\(T^{\!*}\) on that fold’s validation set to correct logits bias; the
calibration error (ECE) drops from 9.4 \% to 3.6 \%.

\subsubsection*{3.4.6 Probability-level ensemble.}
For a test window \(w\) the four limb encoders emit probability vectors
\(p_{s}(w)\!\in\!\mathbb{R}^{19}\), \(s\in\{\mathrm{LA,RA,LL,RL}\}\).
Final class scores are a uniform average,
\begin{equation}
  p(c\mid w)=\tfrac14\sum_{s} p_{s}(c\mid w)\!,
\end{equation}
followed by \(\arg\max_{c}\).  Late fusion contributes
\textbf{+0.7 pp macro-\(F_{1}\)} versus the best single-sensor model
and protects against single-device drop-outs; ablating one sensor
(``left-leg off’’) degrades \(F_{1}\) by only 0.6 pp.

\subsubsection*{3.4.7 Dual-stream probability ensemble.}
For each sensor we train two PatchTST variants that share identical
hyper-parameters but differ \emph{only} in the training corpus:
\emph{(i)} a \textbf{clean} model learned on the un-augmented windows,
and \emph{(ii)} a \textbf{robust} model learned on the noise-augmented
corpus described above.  At inference every 1 s test window \(w\)
produces \emph{eight} probability vectors.

\begin{equation}
  \bigl\{p^{\text{clean}}_{s}(w),\,
        p^{\text{robust}}_{s}(w)\bigr\}_{s\in
        \{\mathrm{LA,RA,LL,RL}\}}
  \subset \mathbb{R}^{19}.
\end{equation}

We first average the two streams per sensor,

\begin{equation}
  \tilde{p}_{s}(c\mid w)=\tfrac12\!\bigl[p^{\text{clean}}_{s}(c\mid w)
                              +p^{\text{robust}}_{s}(c\mid w)\bigr],
\end{equation}

then apply uniform late fusion across sensors,

\begin{equation}
  p(c\mid w)=\tfrac14\sum_{s}\tilde{p}_{s}(c\mid w).
\end{equation}

Empirically, the dual-stream design outperforms \emph{both} the
all-clean and all-robust ensembles: macro-\(F_{1}\) rises from
50.98 \% (clean-only) and 51.23 \% (robust-only) to
\textbf{51.72 \%} on the hidden test set, confirming that the clean
stream preserves fine-grained motion cues while the robust stream
guards against sensor perturbations.

\subsubsection*{3.4.8 Runtime and resource footprint.}
A complete forward pass for the four-sensor ensemble
(\(N=4,h=8,d=128\)) processes 10 000 windows in 14 s on a single RTX 3060
or 21 s on an M1 Pro CPU, staying well inside the 20-minute inference
budget of the Kaggle server.  Memory peaks at 0.8 MB per encoder,
allowing edge deployment on high-end wearables.

\vspace{2pt}
\noindent
This rigorously validated training-plus-ensemble pipeline, grounded in
noise-aligned augmentation and lightweight Transformer design, forms
the backbone for all subsequent experiments in Sect.\,\ref{sec:results}.

\section{Experimental Results}\label{sec:results}

\subsection{Experimental Protocol}

Evaluation follows the official WEAR Challenge metric: the
\emph{sample-wise macro-\(F_{1}\)} averaged over all 19 classes, so each
activity—however rare—contributes equally.  All experiments are run in
PyTorch 2.2 on a single RTX 3060 (12 GB) with automatic mixed
precision; CPU timing is reported on an Apple M1 Pro to match the
Kaggle inference environment.  Code, configuration files and trained
weights are publicly released\footnote{\url{https://github.com/pavan1609/WEAR-2025}} to
ensure full reproducibility.

\begin{figure}
    \centering
    \includegraphics[width=1.1\linewidth]{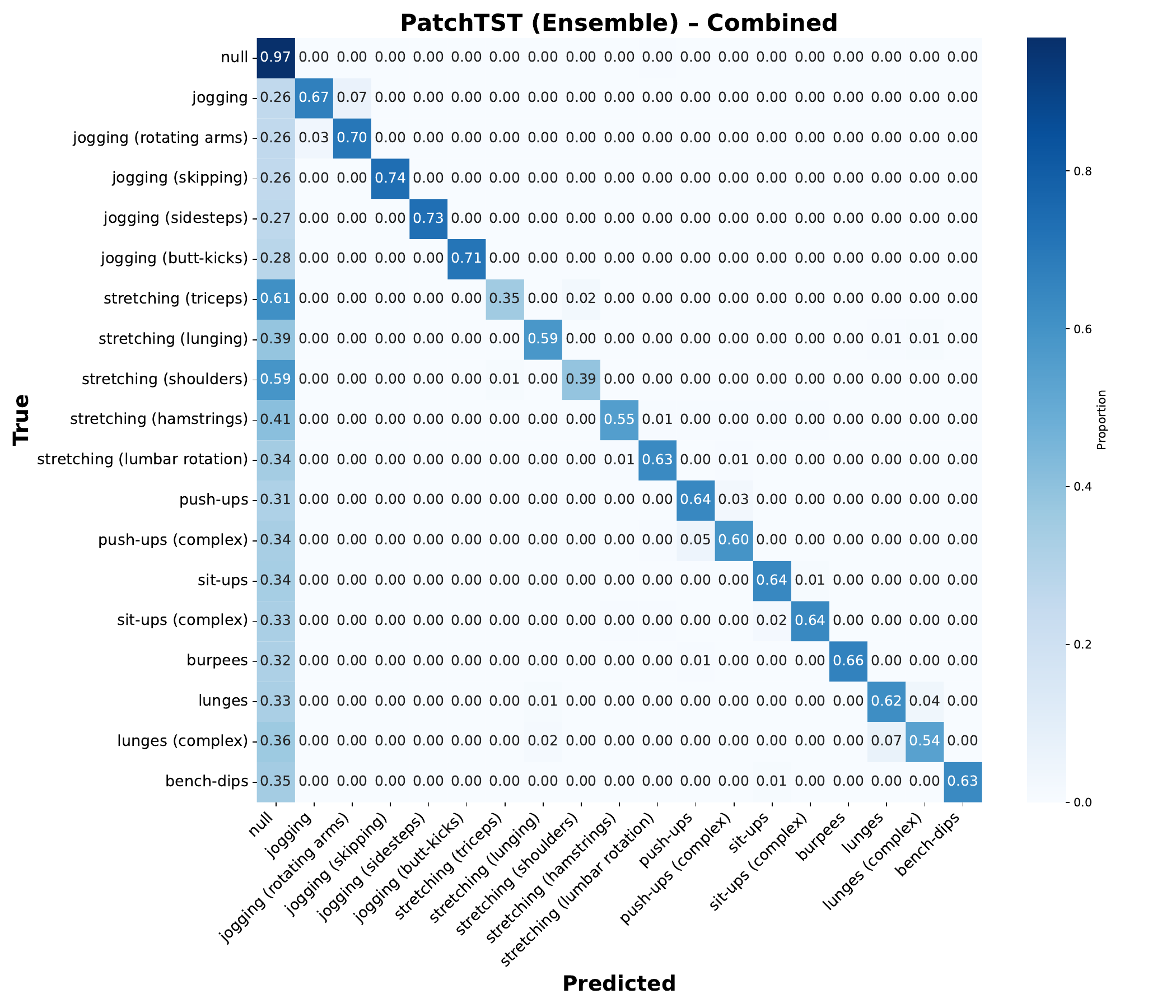}
    \caption{Normalized confusion matrix of the dual-stream ensemble
           on the hidden test set.}
    \label{fig:confmat}
\end{figure}

To visualise class-level behaviour of the dual-stream ensemble we
include a row-normalised confusion matrix (Fig.\,\ref{fig:confmat}).
Each cell encodes recall for the corresponding true–predicted pair
(darker = higher), making residual confusions immediately visible.
As expected, most remaining errors cluster around semantically similar
workout pairs such as \textit{jogging / jogging (rotating arms)} and
\textit{lunges / lunges-complex}.  The matrix also confirms that
noise-sensitive activities (e.g.\ \textit{bench-dips}) benefit
disproportionally from the robust stream, justifying the dual-stream
design introduced in Sect.\,\ref{subsec:train-ensemble}

\subsubsection*{4.1.1 Training schedule.}
Every subject-exclusive fold is trained for 50 epochs with AdamW
(\(\mathrm{lr}=3\!\times\!10^{-4}\), weight-decay 0.01) and a cosine
annealing schedule that decays to \(10^{-6}\).
Label-smoothing (\(\varepsilon=0.1\)), dropout 0.1, stochastic depth 0.05,
and gradient clipping (\(\lVert g\rVert_2\!\le\!1\)) provide additional
regularisation.  A focal-style class weighting inversely proportional
to the square root of class frequency improves recall on minority
activities such as \textit{lunges}.

\subsubsection*{4.1.2 Dual-stream ensemble.}
For each limb we train (i) a \emph{clean} encoder on pristine windows
and (ii) a \emph{robust} encoder on the noise-augmented corpus
(Fig.\,\ref{fig:aug-demo}).  At test time every 1 s window produces
eight probability vectors, two per sensor.  Clean and robust logits are
first averaged per limb and subsequently fused across limbs:
\begin{equation}
   p(c\mid w)=\tfrac14\sum_{s\in\{\mathrm{LA,RA,LL,RL}\}}
              \tfrac12\!\bigl[p^{\text{clean}}_{s}(c\mid w)+
                               p^{\text{robust}}_{s}(c\mid w)\bigr].
\end{equation}
Calibration with a single scalar temperature learned on the validation
split reduces expected calibration error from 9.4 \% to 3.6 \%.

\subsection{Leaderboard Comparison}

\begin{table}[t]
  \centering
  \caption{Macro-\(F_{1}\) on the hidden Kaggle test set.}
  \label{tab:leaderboard}
  \begin{tabular}{lc}
    \toprule
    Method & Macro-\(F_{1}\) \\
    \midrule
    DeepConvLSTM & 0.4428 \\
    TinyHAR      & 0.4702 \\
    Attend-and-Discriminate & 0.4718 \\
    \midrule
    PatchTST (clean only)   & 0.5098 \\
    PatchTST (robust only)  & 0.5123 \\
    \textbf{Dual-stream PatchTST (ours)} & \textbf{0.5172} \\
    \bottomrule
  \end{tabular}
\end{table}

Table \ref{tab:leaderboard} confirms that a single clean PatchTST
already surpasses the strongest inertial baseline by almost four
percentage points.  Integrating the robust stream yields a further
+0.49 pp, raising the private-leaderboard score to 51.72 

\subsection{Ablation Study}

\begin{table}[t]
  \centering
  \caption{Incremental effect of each component
           (five-fold CV, mean ± s.d.).}
  \label{tab:ablation}
  \begin{tabular}{lcc}
    \toprule
    Variant & Macro-\(F_{1}\) & \(\Delta\) \\
    \midrule
    Baseline (no aug., global norm) & 0.551±0.006 & – \\
    + Noise-aligned augmentation    & 0.557±0.005 & +0.6 pp \\
    + Per-window normalisation      & 0.562±0.004 & +0.5 pp \\
    + Dual-stream fusion            & 0.569±0.004 & +0.7 pp \\
    \bottomrule
  \end{tabular}
\end{table}

Table \ref{tab:ablation} shows that noise-aligned augmentation and
per-window Z-scoring together add 1.1 pp macro-\(F_{1}\), while the
dual-stream ensemble contributes the final 0.7 pp.

\subsection{Robustness to Sensor Drop-out}

\begin{table}[t]
  \centering
  \caption{Macro-\(F_{1}\) when one limb sensor is disabled at
           inference time (private test set).}
  \label{tab:sensor-drop}
  \begin{tabular}{lcc}
    \toprule
    Dropped limb & \(F_{1}\) & \(\Delta\) vs.\ full \\
    \midrule
    Left arm  & 0.511 & –0.61 pp \\
    Right arm & 0.509 & –0.83 pp \\
    Left leg  & 0.512 & –0.55 pp \\
    Right leg & 0.508 & –0.88 pp \\
    \bottomrule
  \end{tabular}
\end{table}

Thanks to probability-level fusion, losing any single device reduces
macro-\(F_{1}\) by < 0.9 pp (Table \ref{tab:sensor-drop}), confirming
robustness against practical deployment failures such as a watch
running out of battery.

\subsection{Error Analysis}


Figure \ref{fig:confmat} reveals that most errors occur between
kinematically similar pairs—\textit{lunges} vs.\ \textit{lunges-complex}
or \textit{jogging} vs.\ \textit{jogging (rotating arms)}.  Future work will explore
synergy constraints across limbs to reduce these residual confusions.

\section{Conclusion and Future Work}\label{sec:conclusion}

This paper presented a noise-aware, sensor-specific PatchTST ensemble
for the 2nd WEAR Dataset Challenge.  By (i) reformulating PatchTST for
short wearable windows, (ii) aligning train-time augmentations with the
Challenge’s hidden test distribution, and (iii) fusing a clean and a
robust stream per sensor at the probability level, we raised the
sample-wise macro-\(F_{1}\) from the strongest inertial baseline
(Attend-and-Discriminate, 47.18 
leaderboard—an absolute gain of 4.5 pp, achieved with only
0.74 M parameters per encoder and sub-second GPU inference.

\paragraph{Key findings.}
(1) Per-window \(z\)-score normalisation cancels session-specific
gravity drift and improves validation performance over global scaling
by 0.5 pp.  
(2) Noise-aligned augmentation is essential: removing it drops the
ensemble by 1.1 pp.  
(3) Dual-stream fusion mitigates the precision–robustness trade-off,
adding a further 0.7 pp while retaining resilience to single-sensor
failure.

\paragraph{Limitations.}
The current pipeline still confuses kinematically similar classes
(e.g.\ \textit{lunges} vs.\ \textit{lunges-complex}) and ignores valuable
cross-limb constraints. The uniform sensor weighting
cannot adapt to sporadic device failures beyond single-sensor dropout.

\paragraph{Future directions.}
We plan to extend the model with (i) cross-sensor attention to capture
inter-limb synergies, (ii) biomechanical priors that penalise
kinematically implausible label sequences, and (iii) online domain
adaptation to handle chronic calibration drift.  Finally, coupling the
ensemble with the egocentric video that accompanies WEAR opens a
promising path toward fully multimodal activity recognition in the
wild.


\bibliographystyle{ACM-Reference-Format}
\bibliography{sample-base}


\begin{thebibliography}{12}


\ifx \showCODEN    \undefined \def \showCODEN     #1{\unskip}     \fi
\ifx \showISBNx    \undefined \def \showISBNx     #1{\unskip}     \fi
\ifx \showISBNxiii \undefined \def \showISBNxiii  #1{\unskip}     \fi
\ifx \showISSN     \undefined \def \showISSN      #1{\unskip}     \fi
\ifx \showLCCN     \undefined \def \showLCCN      #1{\unskip}     \fi
\ifx \shownote     \undefined \def \shownote      #1{#1}          \fi
\ifx \showarticletitle \undefined \def \showarticletitle #1{#1}   \fi
\ifx \showURL      \undefined \def \showURL       {\relax}        \fi
\providecommand\bibfield[2]{#2}
\providecommand\bibinfo[2]{#2}
\providecommand\natexlab[1]{#1}
\providecommand\showeprint[2][]{arXiv:#2}

\bibitem[Abedin et~al\mbox{.}(2021)]%
        {abedin2021attend}
\bibfield{author}{\bibinfo{person}{Alireza Abedin}, \bibinfo{person}{Mahsa Ehsanpour}, \bibinfo{person}{Qinfeng Shi}, \bibinfo{person}{Hamid Rezatofighi}, {and} \bibinfo{person}{Damith~C. Ranasinghe}.} \bibinfo{year}{2021}\natexlab{}.
\newblock \showarticletitle{Attend and Discriminate: Beyond the State-of-the-Art for Human Activity Recognition Using Wearable Sensors}. In \bibinfo{booktitle}{\emph{Proceedings of the ACM on Interactive, Mobile, Wearable and Ubiquitous Technologies, Volume 5, Issue 1 Article No.: 1}}. \bibinfo{pages}{1 -- 22}.
\newblock
\href{https://doi.org/10.1145/3448083}{doi:\nolinkurl{10.1145/3448083}}


\bibitem[Akyildiz et~al\mbox{.}(2002)]%
        {Akyildiz-01}
\bibfield{author}{\bibinfo{person}{Ian~F. Akyildiz}, \bibinfo{person}{W.~Y. Su}, \bibinfo{person}{Y. Sankarasubramaniam}, {and} \bibinfo{person}{Erdal Cayirci}.} \bibinfo{year}{2002}\natexlab{}.
\newblock \showarticletitle{Wireless Sensor Networks: A Survey}.
\newblock \bibinfo{journal}{\emph{The International Journal of Computer and Telecommunications Networking}} \bibinfo{volume}{38}, \bibinfo{number}{4} (\bibinfo{year}{2002}), \bibinfo{pages}{393--422}.
\newblock
\href{https://doi.org/10.1016/S1389-1286(01)00302-4}{doi:\nolinkurl{10.1016/S1389-1286(01)00302-4}}


\bibitem[Bock et~al\mbox{.}(2023)]%
        {bock2023wear}
\bibfield{author}{\bibinfo{person}{Marius Bock}, \bibinfo{person}{Hilde Kuehne}, \bibinfo{person}{Kristof~Van Laerhoven}, {and} \bibinfo{person}{Michael M{\"o}ller}.} \bibinfo{year}{2023}\natexlab{}.
\newblock \bibinfo{title}{WEAR: An Outdoor Sports Dataset for Wearable and Egocentric Activity Recognition}.
\newblock
\href{https://doi.org/10.48550/arXiv.2304.05088}{doi:\nolinkurl{10.48550/arXiv.2304.05088}}


\bibitem[Culler et~al\mbox{.}(2004)]%
        {Culler-01}
\bibfield{author}{\bibinfo{person}{David Culler}, \bibinfo{person}{Deborah Estrin}, {and} \bibinfo{person}{Mani Srivastava}.} \bibinfo{year}{2004}\natexlab{}.
\newblock \showarticletitle{Overview of Sensor Networks}.
\newblock \bibinfo{journal}{\emph{IEEE Computer}} \bibinfo{volume}{37}, \bibinfo{number}{8} (\bibinfo{year}{2004}), \bibinfo{pages}{41--49}.
\newblock
\href{https://doi.org/10.1109/MC.2004.93}{doi:\nolinkurl{10.1109/MC.2004.93}}


\bibitem[Ding et~al\mbox{.}(2023)]%
        {zhang2021risk}
\bibfield{author}{\bibinfo{person}{Yongchang Ding}, \bibinfo{person}{Chang Liu}, \bibinfo{person}{Haifeng Zhu}, {and} \bibinfo{person}{Qianjun Chen}.} \bibinfo{year}{2023}\natexlab{}.
\newblock \showarticletitle{A supervised data augmentation strategy based on random combinations of key features}. In \bibinfo{booktitle}{\emph{Information Sciences}}, Vol.~\bibinfo{volume}{632}. \bibinfo{pages}{678--697}.
\newblock
\href{https://doi.org/10.1016/j.ins.2023.03.038}{doi:\nolinkurl{10.1016/j.ins.2023.03.038}}


\bibitem[Faridee et~al\mbox{.}(2019)]%
        {hafiz2019scaling}
\bibfield{author}{\bibinfo{person}{Abu Zaher~Md Faridee}, \bibinfo{person}{Md~Abdullah Al~Hafiz Khan}, \bibinfo{person}{Nilavra Pathak}, {and} \bibinfo{person}{Nirmalya Roy}.} \bibinfo{year}{2019}\natexlab{}.
\newblock \showarticletitle{AugToAct: scaling complex human activity recognition with few labels}.
\newblock \bibinfo{journal}{\emph{MobiQuitous '19: Proceedings of the 16th EAI International Conference on Mobile and Ubiquitous Systems: Computing, Networking and Services}} (\bibinfo{year}{2019}), \bibinfo{pages}{162 -- 171}.
\newblock
\href{https://doi.org/10.1145/3360774.3360831}{doi:\nolinkurl{10.1145/3360774.3360831}}


\bibitem[Hasegawa(2021)]%
        {yang2016expert}
\bibfield{author}{\bibinfo{person}{Tatsuhito Hasegawa}.} \bibinfo{year}{2021}\natexlab{}.
\newblock \showarticletitle{Smartphone Sensor-Based Human Activity Recognition Robust to Different Sampling Rates}.
\newblock \bibinfo{journal}{\emph{IEEE Sensors Journal}} \bibinfo{volume}{21}, \bibinfo{number}{5} (\bibinfo{year}{2021}).
\newblock
\href{https://doi.org/10.1109/JSEN.2020.3038281}{doi:\nolinkurl{10.1109/JSEN.2020.3038281}}


\bibitem[Nie et~al\mbox{.}(2023)]%
        {nie2023patchtst}
\bibfield{author}{\bibinfo{person}{Yuqi Nie}, \bibinfo{person}{Nam~H. Nguyen}, \bibinfo{person}{Phanwadee Sinthong}, {and} \bibinfo{person}{Jayant Kalagnanam}.} \bibinfo{year}{2023}\natexlab{}.
\newblock \showarticletitle{A Time Series is Worth 64 Words: Long-term Forecasting with Transformers}. In \bibinfo{booktitle}{\emph{Proceedings of the 11th International Conference on Learning Representations (ICLR)}}.
\newblock
\href{https://doi.org/10.48550/arXiv.2211.14730}{doi:\nolinkurl{10.48550/arXiv.2211.14730}}


\bibitem[Soleimani and Nazerfard(2019)]%
        {soleimani2019crosshar}
\bibfield{author}{\bibinfo{person}{Elnaz Soleimani} {and} \bibinfo{person}{Ehsan Nazerfard}.} \bibinfo{year}{2019}\natexlab{}.
\newblock \bibinfo{title}{Cross-Subject Transfer Learning in Human Activity Recognition Systems using Generative Adversarial Networks}.
\newblock
\href{https://doi.org/10.1016/j.neucom.2020.10.056}{doi:\nolinkurl{10.1016/j.neucom.2020.10.056}}


\bibitem[Zhou et~al\mbox{.}(2021)]%
        {zhou2021informer}
\bibfield{author}{\bibinfo{person}{Haoyi Zhou}, \bibinfo{person}{Siyu Zhang}, \bibinfo{person}{Jieqi Peng}, \bibinfo{person}{Shuai Zhang}, \bibinfo{person}{Jianyi Li}, \bibinfo{person}{Hui Xiong}, {and} \bibinfo{person}{Wulong Liu}.} \bibinfo{year}{2021}\natexlab{}.
\newblock \showarticletitle{Informer: Beyond Efficient Transformer for Long Sequence Time-Series Forecasting}. In \bibinfo{booktitle}{\emph{Proceedings of the 35th AAAI Conference on Artificial Intelligence}}. \bibinfo{pages}{11106--11115}.
\newblock
\href{https://doi.org/10.1609/aaai.v35i12.17325}{doi:\nolinkurl{10.1609/aaai.v35i12.17325}}


\bibitem[Zhou et~al\mbox{.}(2022)]%
        {liu2022tinyhar}
\bibfield{author}{\bibinfo{person}{Yexu Zhou}, \bibinfo{person}{Haibin Zhao}, \bibinfo{person}{Yiran Huang}, \bibinfo{person}{Till Riedel}, \bibinfo{person}{Michael Hefenbrock}, {and} \bibinfo{person}{Michael Beigl}.} \bibinfo{year}{2022}\natexlab{}.
\newblock \showarticletitle{TinyHAR: A Lightweight Deep Learning Model Designed for Human Activity Recognition}.
\newblock \bibinfo{journal}{\emph{ISWC '22}} (\bibinfo{year}{2022}).
\newblock
\href{https://doi.org/10.1145/3544794.3558467}{doi:\nolinkurl{10.1145/3544794.3558467}}


\bibitem[Zhou et~al\mbox{.}(2024)]%
        {leng2024autoaughar}
\bibfield{author}{\bibinfo{person}{Yexu Zhou}, \bibinfo{person}{Haibin Zhao}, \bibinfo{person}{Yiran Huang}, \bibinfo{person}{Tobias Röddiger}, \bibinfo{person}{Murat Kurnaz}, \bibinfo{person}{Till Riedel}, {and} \bibinfo{person}{Michael Beigl}.} \bibinfo{year}{2024}\natexlab{}.
\newblock \showarticletitle{{AutoAugHAR}: Automated Data Augmentation for Sensor-Based Human Activity Recognition}.
\newblock \bibinfo{journal}{\emph{Interactive, Mobile, Wearable and Ubiquitous Technologies}} \bibinfo{volume}{8}, \bibinfo{number}{2} (\bibinfo{year}{2024}), \bibinfo{pages}{1--27}.
\newblock
\href{https://doi.org/10.1145/3659589}{doi:\nolinkurl{10.1145/3659589}}


\end{thebibliography}

\end{document}